\documentclass[11pt,a4paper]{article}
\pdfoutput=1
\usepackage[hyperref]{acl2019}
\usepackage[pdftex]{graphicx}
\usepackage[T1]{fontenc}
\usepackage[utf8]{inputenc}
\usepackage{times}
\usepackage{latexsym}
\usepackage{MnSymbol}
\usepackage{amsmath}
\usepackage[capitalize]{cleveref}
\usepackage{amsfonts}
\usepackage{url}
\usepackage{color,soul}
\usepackage{bm}
\usepackage{multirow}
\usepackage[font=small]{caption}
\usepackage{subcaption}
\usepackage[noend]{algpseudocode}
\usepackage{afterpage}
\usepackage{tabularx}
\usepackage{booktabs}
\usepackage{arydshln}
\usepackage{url}
\usepackage[shortlabels]{enumitem}
\usepackage{bigstrut}
\usepackage{array}
\usepackage{microtype}
\usepackage{pifont}
\usepackage{flushend}

\graphicspath{{img/}}

\usepackage{floatrow}
\setkeys{Gin}{width=\linewidth}

\definecolor{mypink1}{rgb}{0.858, 0.188, 0.478}
\definecolor{maroon}{RGB}{220, 20, 60}
\definecolor{green}{RGB}{0, 100, 0}

\aclfinalcopy %
\def\aclpaperid{2160} %

\title{Towards Multimodal Sarcasm Detection \\ (An  \textit{Obviously} Perfect Paper)}

\author{Santiago Castro\(^\dagger\)\thanks{\ \ Equal contribution.} \ , Devamanyu Hazarika\(^\Phi\)\footnotemark[1]  \ , Verónica Pérez-Rosas\(^\dagger\),\\ \textbf{Roger Zimmermann\(^\Phi\),  Rada Mihalcea\(^\dagger\), Soujanya Poria\(^\iota\)}\\
\(^\dagger\)Computer Science \& Engineering, University of Michigan, USA\\
\(^\Phi\)School of Computing, National University of Singapore, Singapore \\
\(^\iota\)Information Systems Technology and Design, SUTD, Singapore \\
\texttt{\{sacastro,vrncapr,mihalcea\}@umich.edu}, \\ \texttt{\{hazarika,rogerz\}@comp.nus.edu.sg}, \texttt{sporia@sutd.edu.sg}}

\begin{document}

\maketitle

\begin{abstract}

Sarcasm is often expressed through several verbal and non-verbal cues, e.g., a change of tone, overemphasis in a word, a drawn-out syllable, or a straight looking face. %
Most of the recent work in sarcasm detection has been carried out on textual data. In this paper, we argue that incorporating multimodal cues can improve the automatic classification of sarcasm. As a first step towards enabling the development of multimodal approaches for sarcasm detection, we propose a new sarcasm dataset, Multimodal Sarcasm Detection Dataset (MUStARD\footnote{MUStARD is an abbreviation for {\bf MU}ltimodal {\bf SAR}casm {\bf D}ataset. Similar to how ``mustard'' adds spice to our food and meals, we believe sarcasm adds spice to our interactions and lives. }), compiled from popular TV shows. MUStARD consists of audiovisual utterances annotated with sarcasm labels. Each utterance is accompanied by its context of historical utterances in the dialogue, which provides additional information on the scenario where the utterance occurs. %
Our initial results show that the use of multimodal information can reduce the relative error rate of sarcasm detection by up to 12.9\% in F-score when compared to the use of individual modalities. The full dataset is publicly available for use at \url{https://github.com/soujanyaporia/MUStARD}.
\end{abstract}

\section{Introduction}

\begin{figure*}[t]
	\includegraphics[width=0.9\linewidth]{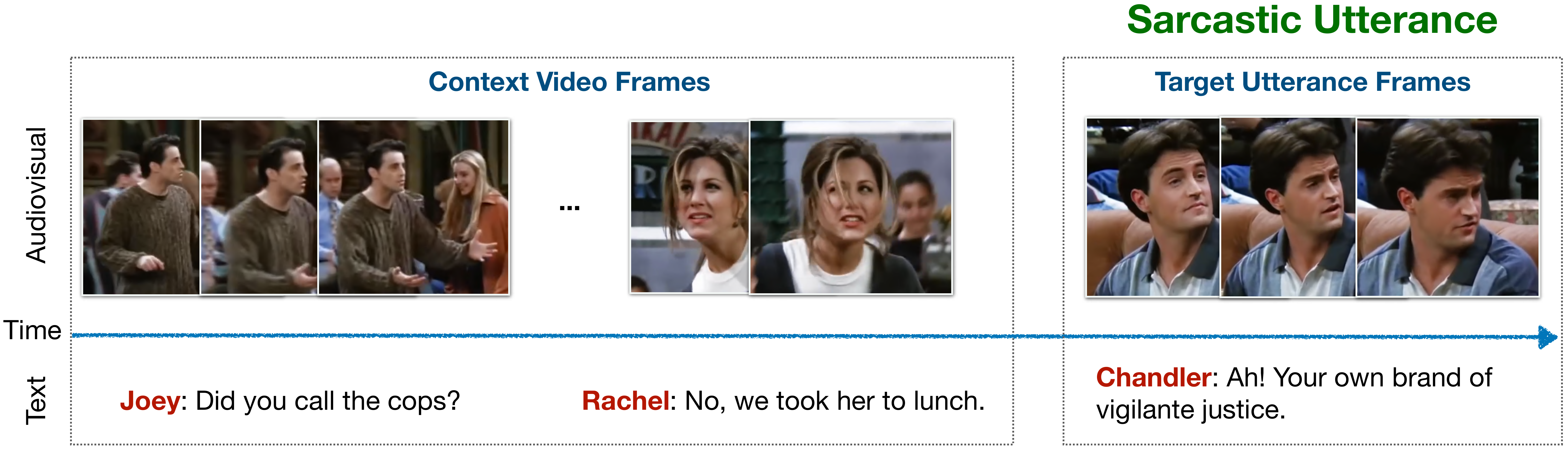}
	\caption{Sample sarcastic utterance in the dataset along with its context and transcript.}
	\label{fig:dataset_instance}
\end{figure*}

Sarcasm plays an important role in daily conversations by allowing individuals to express their intent to mock or display contempt. It is achieved by using irony that reflects a negative connotation. For example, in the utterance:
\textit{Maybe it's a good thing we came here. It's like a lesson in what not to do}, the sarcasm is explicit as the speaker expresses learning of a lesson in a positive light when in reality, she means it in a negative way. However, there are also scenarios where sarcasm lacks explicit linguistic markers, thus requiring additional cues that can reveal the speaker's intentions.  For instance, sarcasm can be expressed using a combination of verbal and non-verbal cues, such as a change of tone, overemphasis in a word, a drawn-out syllable, or a straight looking face. %
Moreover, sarcasm detection involves finding linguistic or contextual incongruity, which in turn requires further information, either from multiple modalities~\cite{schifanella2016detecting,mishra2016predicting} or from the context history in a dialogue.%

This paper explores the role of multimodality and conversational context in sarcasm detection and introduces a new resource to further enable research in this area.
More specifically, our paper makes the following contributions: (1) We curate a new dataset, MUStARD, for multimodal sarcasm research with high-quality annotations, including both mutlimodal and conversational context features; (2) We exemplify various scenarios where incongruity in sarcasm is evident across different modalities, thus stressing the role of multimodal approaches to solve this problem; (3) We introduce several baselines and show that multimodal models are significantly more effective when compared to their unimodal variants; and %
(4) We also provide preceding turns in the dialogue which act as context information. Consequently, we surmise that this property of MUStARD leads to a new sub-task for future work: \textit{sarcasm detection in conversational context.}

The rest of the paper is organized as follows. \Cref{sec:related_work} summarizes previous work on sarcasm detection using both unimodal and multimodal sources. \Cref{sec:dataset} describes the dataset collection, the annotation process, and the types of sarcastic situations covered by our dataset. \Cref{sec:feature_extraction} explains how we extract features for the different modalities. \Cref{sec:experiments} shows the experimental work around the new dataset while \cref{sec:multimodal_prediction} analyzes it. Finally, \Cref{sec:conclusion} offers  conclusions and discusses open problems related to this resource.

\section{Related Work}%
\label{sec:related_work}

Automated sarcasm detection has gained increased interest in recent years. It is a widely studied linguistic device whose significance is seen in sentiment analysis and human-machine interaction research.  Various research projects have approached this problem through different modalities, such as text, speech, and visual data streams.

\paragraph{Sarcasm in Text:}

Traditional approaches for detecting sarcasm in text have considered rule-based techniques~\cite{veale2010detecting}, lexical and pragmatic features~\cite{carvalho2009clues}, stylistic features~\cite{davidov2010semi}, situational disparity~\cite{riloff2013sarcasm}, incongruity~\cite{joshi2015harnessing}, or user-provided annotations such as hashtags~\cite{liebrecht2013perfect}.

Resources in this domain are collected using Twitter as a primary data source and are annotated using two main strategies: manual annotation~\cite{riloff2013sarcasm,joshi2016cultural} and distant supervision through hashtags~\cite{davidov2010semi,abercrombie2016putting}. %
Other research leverages context to acquire shared knowledge between the speaker and the audience~\cite{wallace2014humans, bamman2015contextualized}. A variety of contextual features have been explored, including speaker's background and behavior in online platforms~\cite{rajadesingan2015sarcasm}, embeddings of expressed sentiment and speaker's personality traits~\cite{poria2016deeper}, learning of user-specific representations~\cite{wallace2016modelling, kolchinski2018representing}, user-community features~\cite{wallace2015sparse}, as well as stylistic and discourse features~\cite{hazarika2018cascade}. In our dataset, we capitalize on the conversational format and provide context by including preceding utterances along with speaker identities. To the best of our knowledge, there is no prior work which deals with the task of sarcasm detection in conversation.

\paragraph{Sarcasm in Speech:} Sarcasm detection in speech has mainly focused on the identification of prosodic cues in the form of acoustic patterns that are related to sarcastic behavior. Studied features include mean amplitude, amplitude range, speech rate, harmonics-to-noise ratio, and others~\cite{cheang2008sound}. \citet{rockwell2000lower} presented one of the initial approaches to this problem that studied the vocal tonalities of sarcastic speech. They found slower speaking rates and greater intensity as probable markers for sarcasm. Later, \citet{tepperman2006yeah} studied prosodic and spectral features of sound --- both in and out of context --- to determine sarcasm. In general, prosodic features such as intonation and stress are considered important indicators of sarcasm~\cite{bryant2010prosodic,woodland2011context}. We take motivation from this previous research and include similar speech parameters as features in our dataset and baseline experiments.

\paragraph{Multimodal Sarcasm:} Contextual information for sarcasm in text can be included from other modalities. These modalities help in providing additional cues in the form of both common or contrasting patterns. Prior work mainly considers multimodal learning for the readers' ability to perceive sarcasm. Such research couples textual features with cognitive features such as the gaze-behavior of readers~\cite{mishra2016predicting, mishra2016harnessing, mishra2017learning} or electro/magneto-encephalographic (EEG/MEG) signals~\cite{filik2014testing, thompson2016emotional}. In contrast, there is limited work exploring multimodal avenues to understand sarcasm conveyed by the opinion holder. \citet{attardo2003multimodal} presented one of the preliminary explorations on this topic where different phonological and visual markers for sarcasm were studied. However, this work did not analyze the interplay of the modalities. More recently, \citet{schifanella2016detecting} presented a multimodal approach for this task by considering visual content accompanying text in online sarcastic posts. They extracted semantic visual features from images using pre-trained networks and fused them with textual features. In our work, we extend these notions and propose to analyze video-based sarcasm in dialogues. To the best of our knowledge, ours is the first work to propose a resource on video-level sarcasm. \citet{joshi2016harnessing} proposed a dataset similar to us, i.e., based on the TV show \textit{Friends}. However, their corpus  only includes the textual modality and is thus not multimodal in nature. Furthermore, we also analyze multiple challenges in sarcasm that call for multimodal learning and provide an evaluation setup for future works to test upon.

\section{Dataset}%
\label{sec:dataset}

To enable the exploration of multimodal sarcasm detection, we introduce a new dataset (MUStARD) consisting of short videos manually annotated for their sarcasm property. 
\subsection{Data Collection}

To collect potentially sarcastic examples, we conduct web searches on differences sources, mainly YouTube. We use keywords such as \textit{Friends sarcasm}, \textit{Chandler sarcasm}, \textit{Sarcasm 101}, \textit{Sarcasm in TV shows}. Using this strategy, we obtain videos from three main TV shows: Friends, The Golden Girls, and Sarcasmaholics  Anonymous. Note that during this initial search, we focus exclusively on sarcastic content. To obtain non-sarcastic videos, we select a subset of 400 videos from MELD, a multimodal emotion recognition dataset derived from the Friends TV series, originally collected by~\citet{poria2018meld}. %
In addition, we collect videos from The Big Bang Theory, a TV show whose characters are often perceived as sarcastic. We obtain videos from seasons 1--8 and segment episodes using laughter cues from its audience. %
Specifically, we use open-source software for laughter detection~\cite{ryokai2018capturing} to obtain initial segmentation boundaries and fine-tune them using the subtitles' timestamps.

The collected set consists of 6,421 videos. Note that although some of the videos in our initial pool include information about their sarcastic nature, the majority of our videos are not labeled. Thus, we conduct a manual annotation as described next.

\subsection{Annotation Process}
\begin{figure}[t]
	\includegraphics[width=0.9\linewidth]{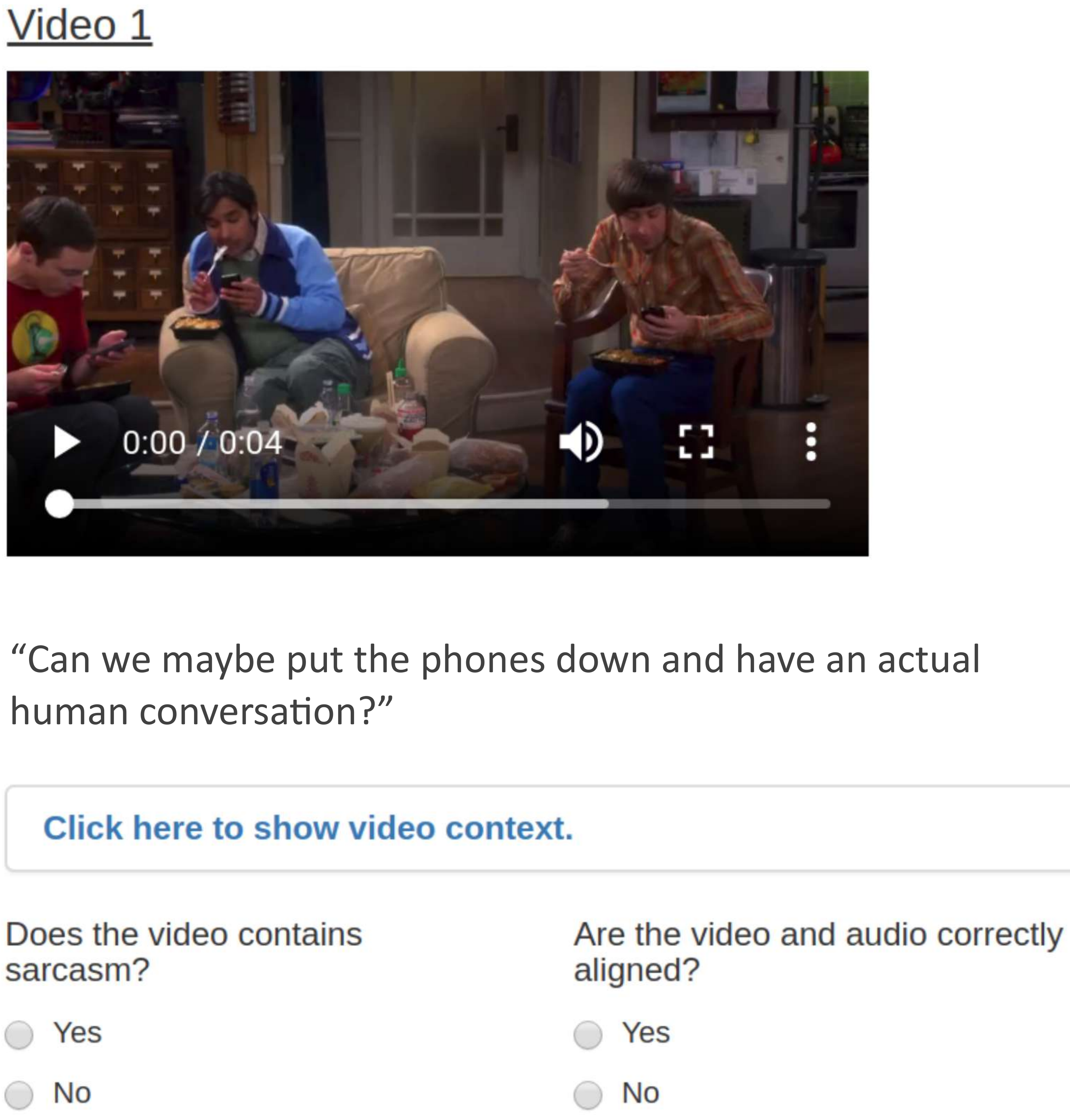}
	\caption{Graphical user interface used by the annotators to label the videos in our dataset.}
	\label{fig:website}
\end{figure}

We built a web-based annotation interface that shows each video along with its transcript and requests annotations for sarcasm. We also ask the annotators to flag misaligned videos, i.e., cases where the audio or video is not properly synchronized. The interface allows the annotators to watch a context video consisting of the previous video utterances, whenever they deem it necessary. Given the large number of videos to be annotated, we request annotations in batches of four videos at a time. Our web interface is shown in \cref{fig:website}.

We conduct the annotation in two steps. First, we annotate the videos from The Big Bang Theory, as it contains the largest set of videos. Second, we annotate the remaining videos, belonging to the other sources.
The annotation is conducted by two graduate students who have first been provided with easy examples of explicit sarcastic content, to illustrate sarcasm in videos. Each annotator labeled the full set of videos independently.

For the first step, after annotating the first part -- consisting of 5,884 utterances from The Big Bang Theory -- we noticed that the majority of them were labeled as non-sarcastic (98\% were considered as non-sarcastic by both). In addition, our initial inter-annotation agreement was low (Kappa score is 0.1463). We thus decided to stop the annotation process and reconcile the annotation differences before proceeding further. The annotators discussed their disagreements for a subset of 20 videos, and then re-annotated the videos. 
This time, we obtained an improved inter-annotator agreement of 0.2326. The annotation disagreements were reconciled by a third annotator by identifying the disagreement cases, watching the videos again and deciding which is the correct label for each one.

Next, we annotate the second part, consisting of 624 videos drawn from Friends, The Golden Girls, and Sarcasmaholics Anonymous. As before, the two annotators label each video independently. The inter-annotator agreement was calculated with a Kappa score of 0.5877. Again, the differences were reconciled by a third annotator.

The resulting set of annotations consists of 345 videos labeled as sarcastic and 6,020 videos labeled as non-sarcastic for a total of 6,365 videos.

\subsection{Transcriptions}

Since we collect videos from several sources, some of them had subtitles or transcripts readily available. This is particularly the case for videos from Big Bang Theory and MELD. We use the MELD transcriptions directly. For Big Bang Theory, we extracted the transcript by applying manual sub-string matching on the episode subtitles. The remaining videos are manually transcribed.

\subsection{Sarcasm Dataset: MUStARD} \label{sec:dataset_details}

\begin{figure}[b]
    \begin{subfigure}{.5\textwidth}
        \centering
        \includegraphics[width=.8\linewidth]{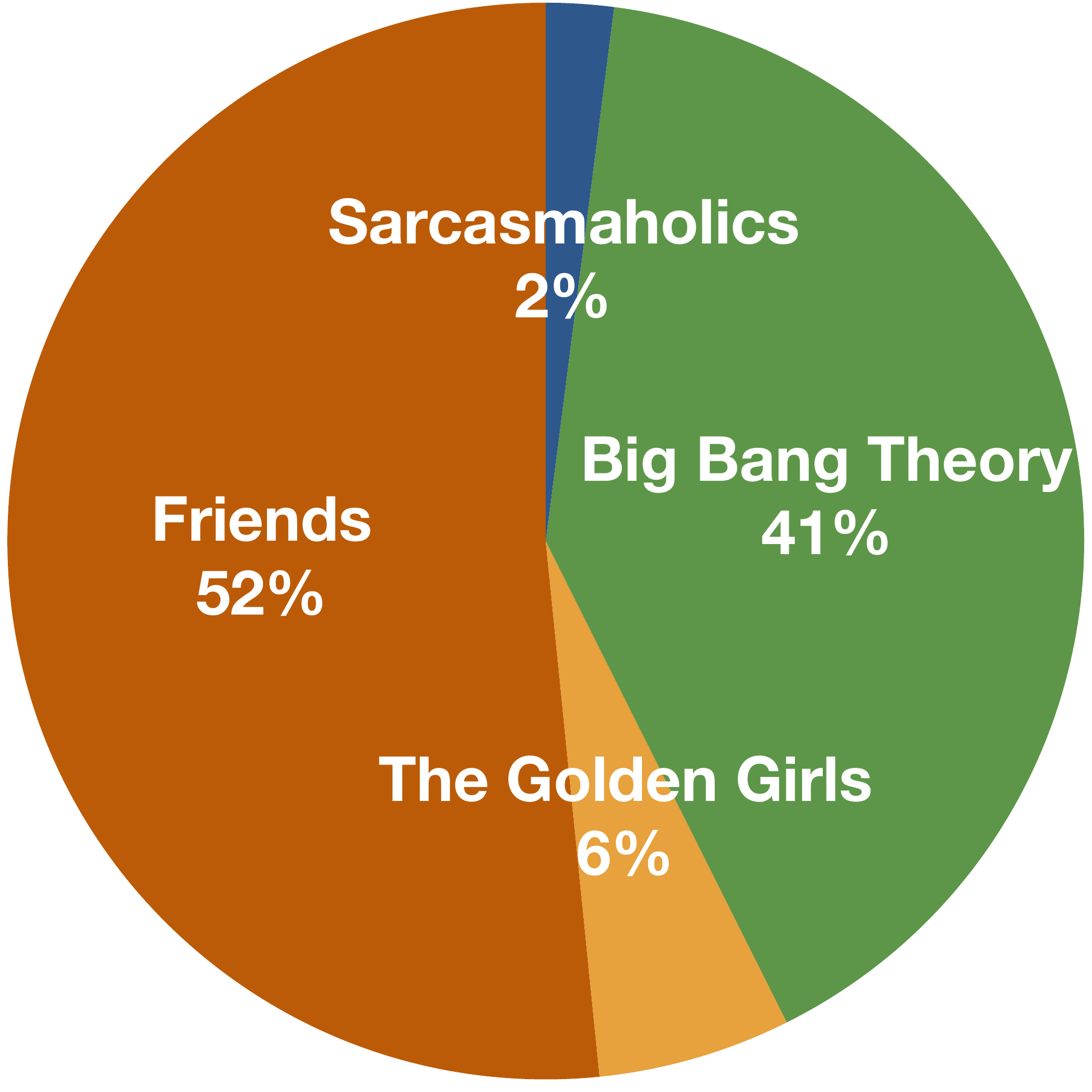}
        \caption{Source across the dataset.}
        \label{fig:statistics1}
    \end{subfigure}%
    \begin{subfigure}{.5\textwidth}
        \centering
        \includegraphics[width=.95\linewidth]{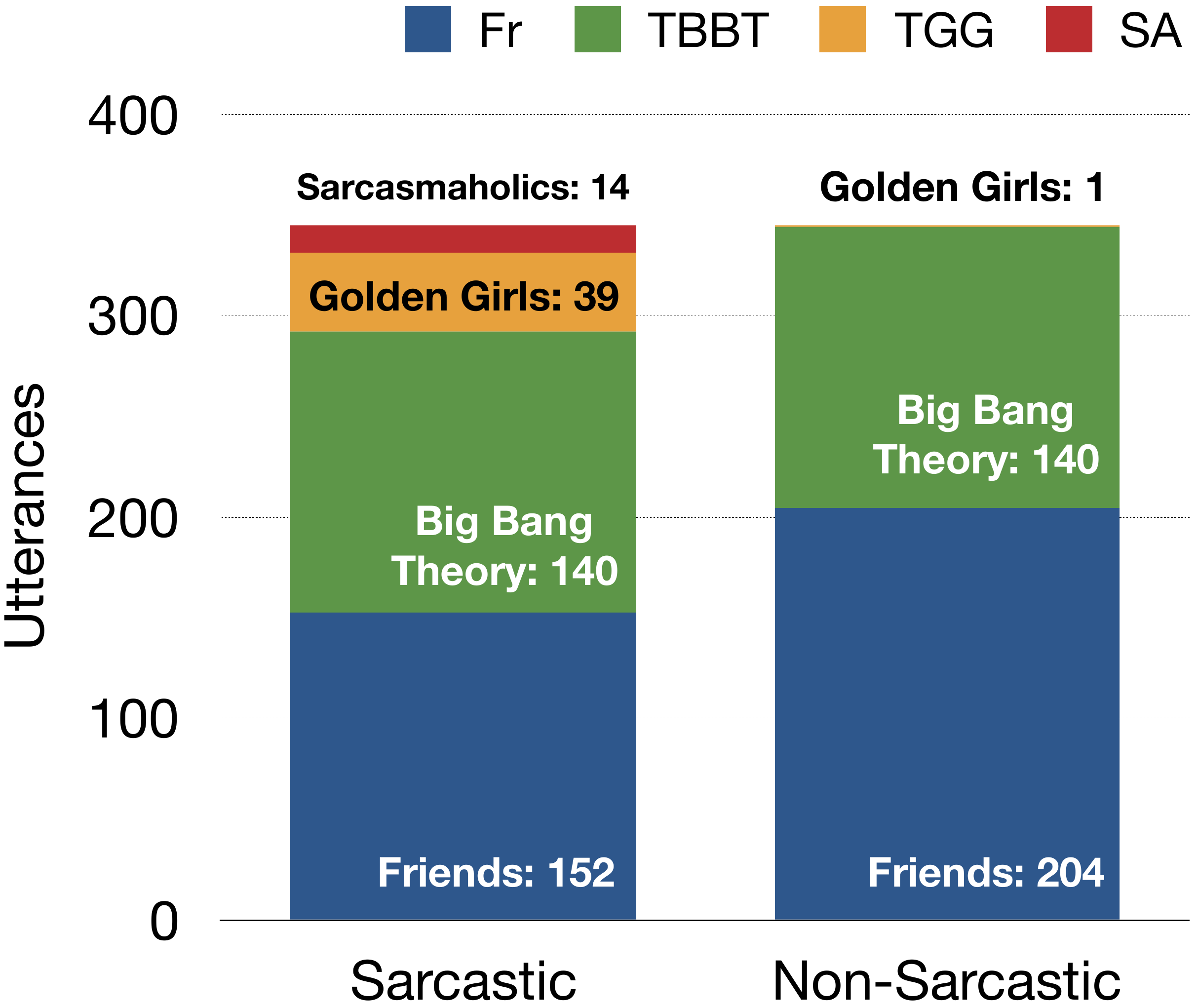}
        \caption{Labels across source shows.}
        \label{fig:statistics2}
    \end{subfigure}%
    \caption{Ratio of the TV shows composing the dataset.}
    \label{fig:statistics}
\end{figure}

\begin{figure*}[t]
    \resizebox{\linewidth}{!}{%
        \begin{subfigure}{.60\textwidth}
          \centering
          \includegraphics[width=\linewidth]{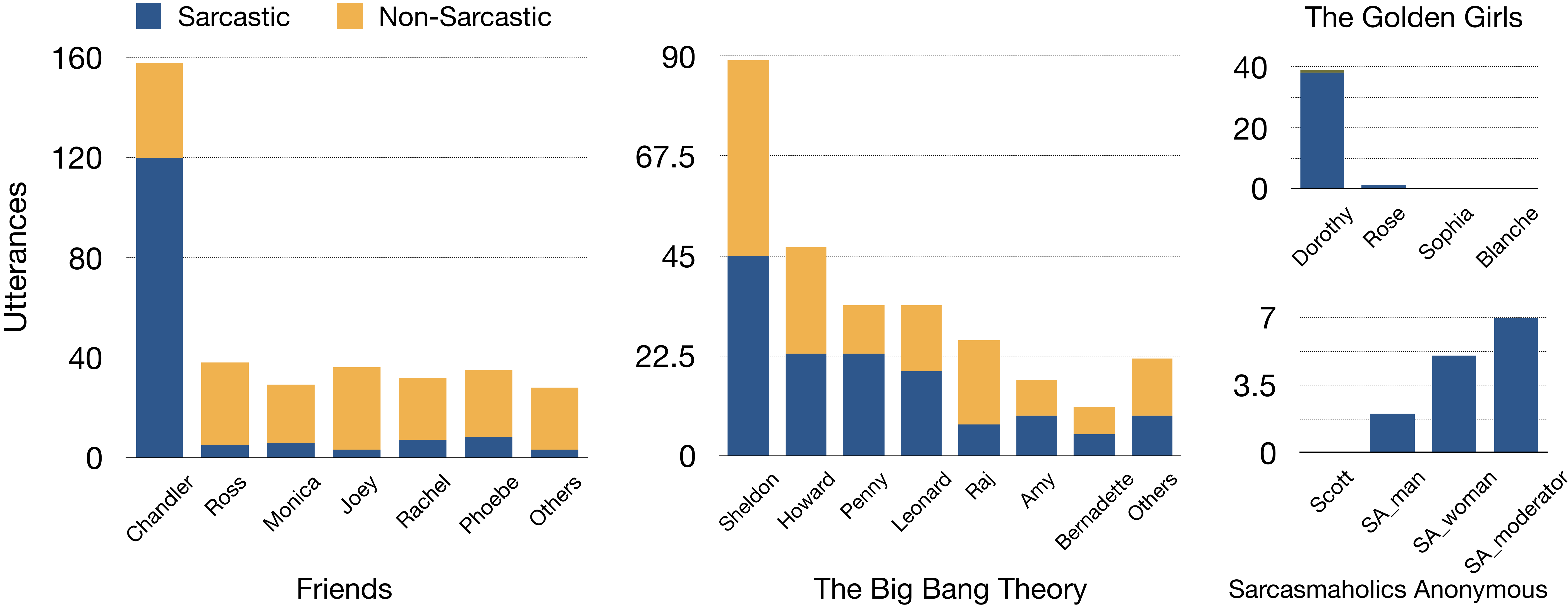}
          \caption{Character-label ratio per source.}
          \label{fig:characters1}
        \end{subfigure}%
        \begin{subfigure}{.33\textwidth}
          \centering
          \includegraphics[width=.85\linewidth]{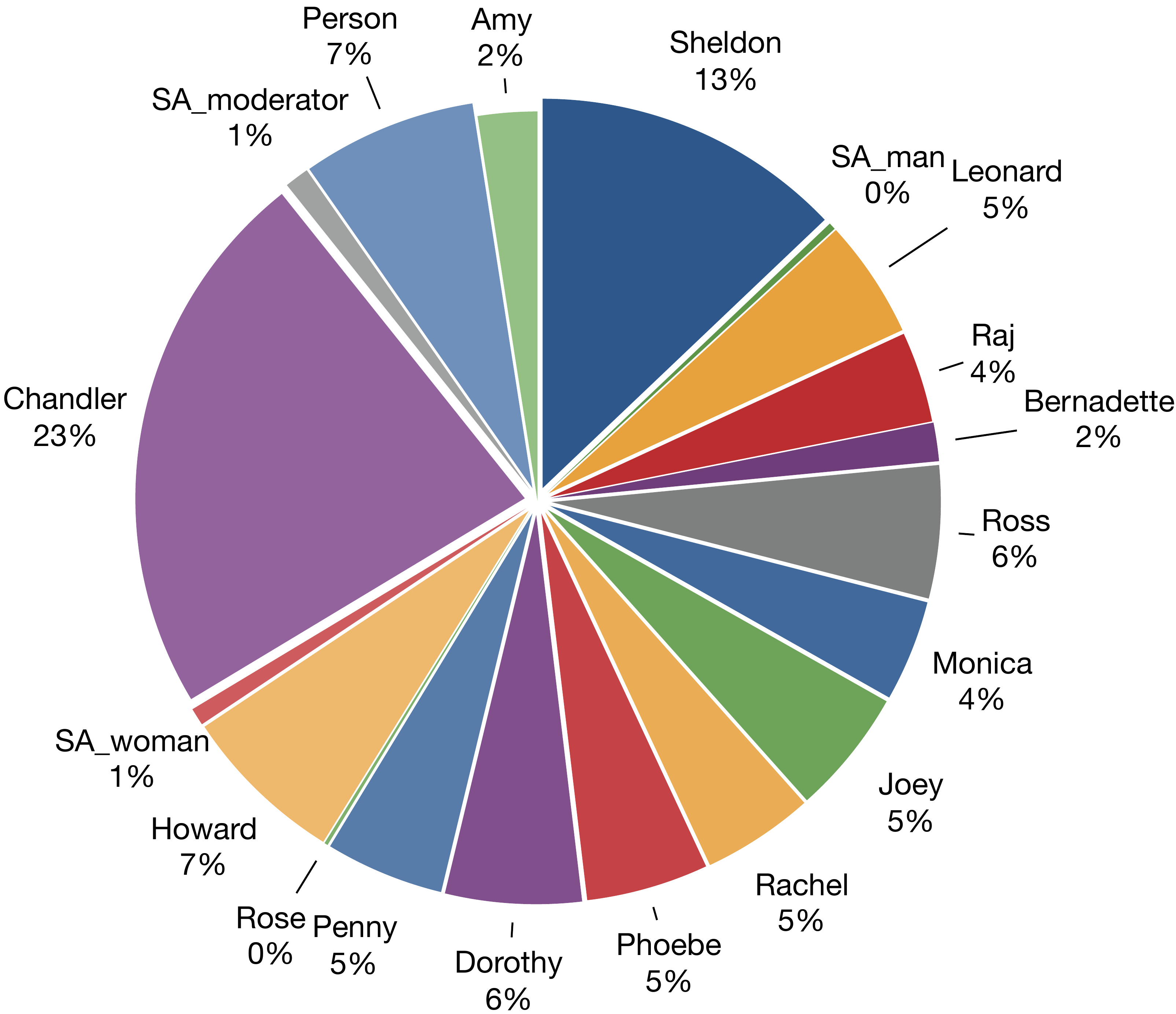}
          \caption{Overall character distribution.}
          \label{fig:characters2}
        \end{subfigure}%
    }
    \caption{Speaker statistics}
    \label{fig:characters}
\end{figure*}

To enable our experiments, which focus explicitly on the multimodal aspects of sarcasm, we decided to work with a balanced sample of sarcastic and non-sarcastic videos. We thus obtain a balanced sample from the set of 6,365 annotated videos. We start by selecting all videos marked as sarcastic from the full set, and then we randomly obtain an equally sized non-sarcastic sample from the non-sarcastic subset by prioritizing the ones annotated by a larger number of annotators. Our dataset thus comprises 690 videos with an even number of sarcastic and non-sarcastic labels. Source, character, and label-ratio statistics are shown in \cref{fig:statistics,fig:characters}.

In the remainder of this paper, we use the term utterance while referring to the videos in our dataset. We extend the definition of an utterance\footnote{An utterance is usually defined as a unit of speech bounded by breaths or pauses.} to include consecutive multi-sentence dialogues of the same speaker to prioritize completeness of information. As a result, $61.3\%$ of the utterances from the dataset are single sentences, while the remaining utterances consist of two or more sentences. %
Each utterance in our dataset is coupled with its context utterances, which are preceding turns by the speakers participating in the dialogue. Some of the context videos contain multi-party dialogue between speakers participating in the scene. The number of turns in the context is manually set to include a coherent background of the target utterance. \cref{tab:datasetstat} shows general statistics for the utterances in our dataset. 

\begin{table}[t]
\small
    \resizebox{\linewidth}{!}{%
	\begin{tabular}{lcc}
		{Statistics} & {Utterance} & {Context} \\
		\midrule
		{Unique words}                 & 1991 & 3205\\
		{Avg. utterance length (tokens)}     & 14%
		& 10%
		\\
		{Max. utterance length (tokens)}              & 73 & 71 \\
		{Avg. duration (seconds)}             & 5.22 & 13.95 \\
		\bottomrule
	\end{tabular}
	}
	\caption{ Dataset statistics by utterance and context. %
	}
	\label{tab:datasetstat}
\end{table}

Each utterance and its context consists of three modalities: video, audio, and transcription (text). Also, all the utterances are accompanied by their speaker identifiers. \cref{fig:dataset_instance} illustrates a sarcastic utterance along with its associated context in the dataset. \cref{fig:characters2} provides the list of major characters present in the dataset. \cref{fig:characters1} details the distribution of labels per character.
Some of the characters, such as \textit{Chandler} and \textit{Sheldon}, occupy major portions of the dataset. This is expected since they play comic roles in the shows. To avoid speaker bias of such popular characters, we also include non-sarcastic samples for these characters. In contrast, the dataset intentionally includes minor roles such as \textit{Dorothy} from The Golden Girls, who is entirely sarcastic throughout the corpus. This allows the study of speaker bias for sarcasm detection.

\begin{figure}[t]
	\includegraphics[width=0.9\linewidth]{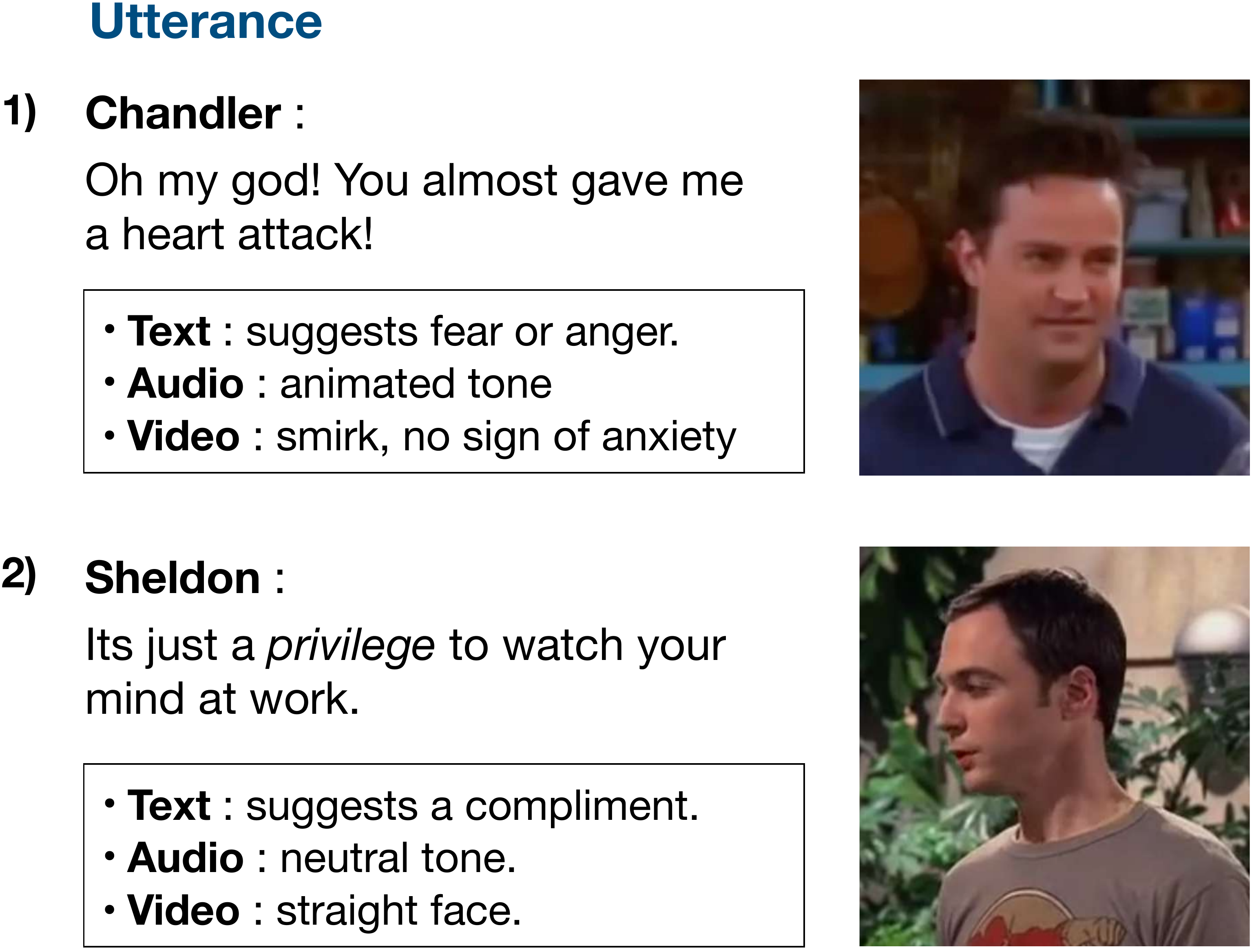}
	\caption{Incongruent modalities in sarcasm.}
	\label{fig:multimodal_examples}
\end{figure}

\subsection{Qualitative Aspects} \label{sec:qualitative}

Sarcasm detection in text often requires additional information that can be leveraged from associated modalities. Below, we analyze some cases that require multimodal reasoning. We exemplify using instances from our proposed dataset to further support our claim of sarcasm being often expressed in a multimodal way.

\paragraph{Role of Multimodality:}

\cref{fig:multimodal_examples} presents two cases where sarcasm is expressed through the incongruity between modalities. In the first case, the language modality indicates fear or anger, whereas the facial modality lacks any visible sign of anxiety that would corroborate the textual modality. In the second case, the text is indicative of a compliment, but the vocal tonality and facial expressions show indifference. In both cases, there exists incongruity between modalities, which acts as a strong indicator of sarcasm.

\begin{figure}[t]
	\includegraphics[width=0.9\linewidth]{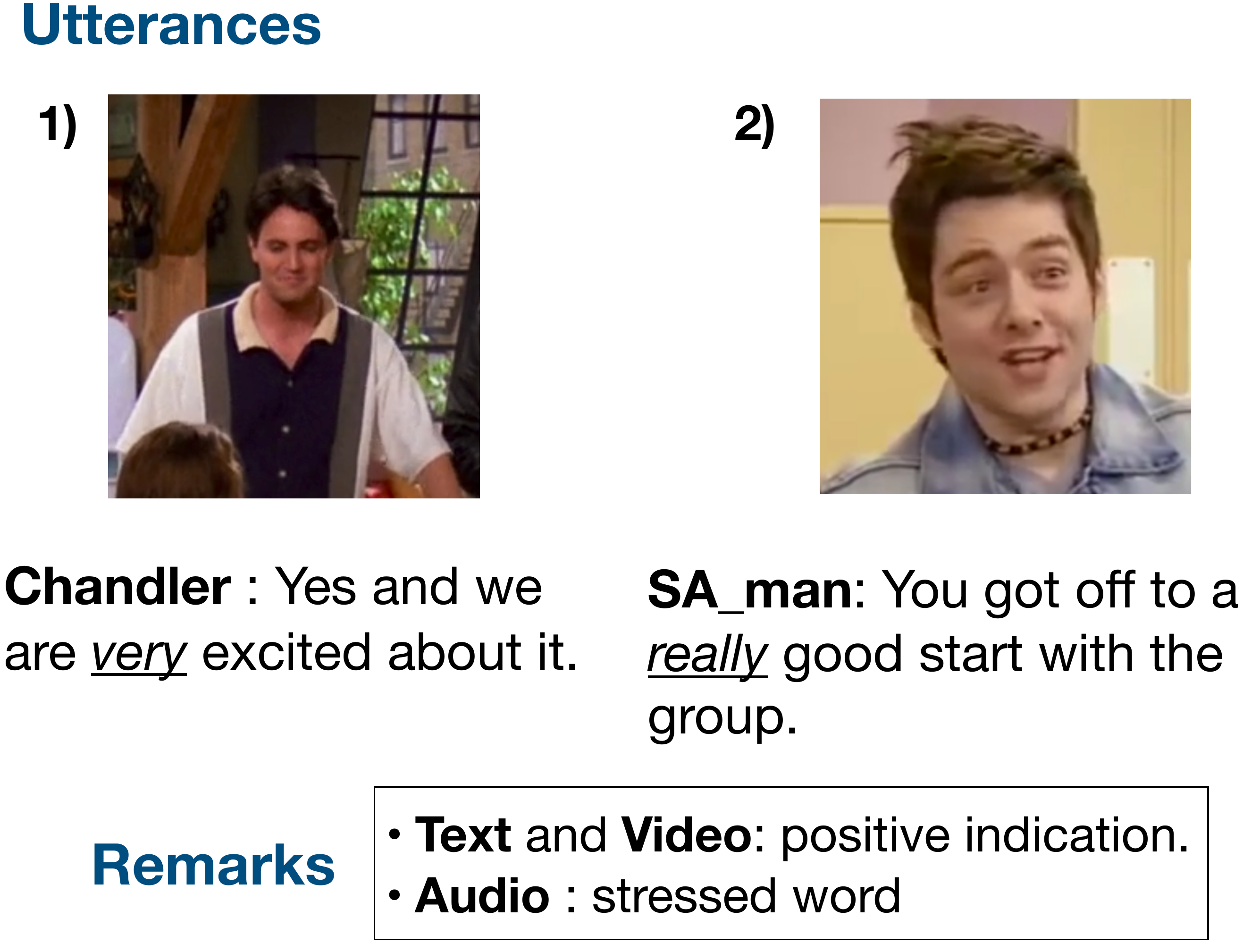}
	\caption{Vocal stress in sarcasm.}
	\label{fig:vocal_stressors}
\end{figure}

Multimodal information is also important in providing additional cues for sarcasm. For example, the vocal tonality of the speaker often indicates sarcasm. Text that otherwise looks seemingly straightforward is noticed to contain sarcasm only when the associated voices are heard. Sarcastic tonalities can range from self-deprecatory or broody tone to something obnoxious and raging. Such extremities are often seen while expressing sarcasm. Another marker of sarcasm is the undue stress on particular words. %
For instance, in the phrase \textit{You did ``really'' well}, if the speaker stresses the word \textit{really}, then the sarcasm is evident. \cref{fig:vocal_stressors} provides sarcastic cases from the dataset where such vocal stresses exist.

It is important to note that sarcasm does not necessarily imply conflicting modalities. Rather, the availability of complementary information through multiple modalities improves the capacity of models to learn discriminative patterns responsible for this complex process.

\paragraph{Role of Context:}

\begin{figure}[t]
	\includegraphics[width=0.9\linewidth]{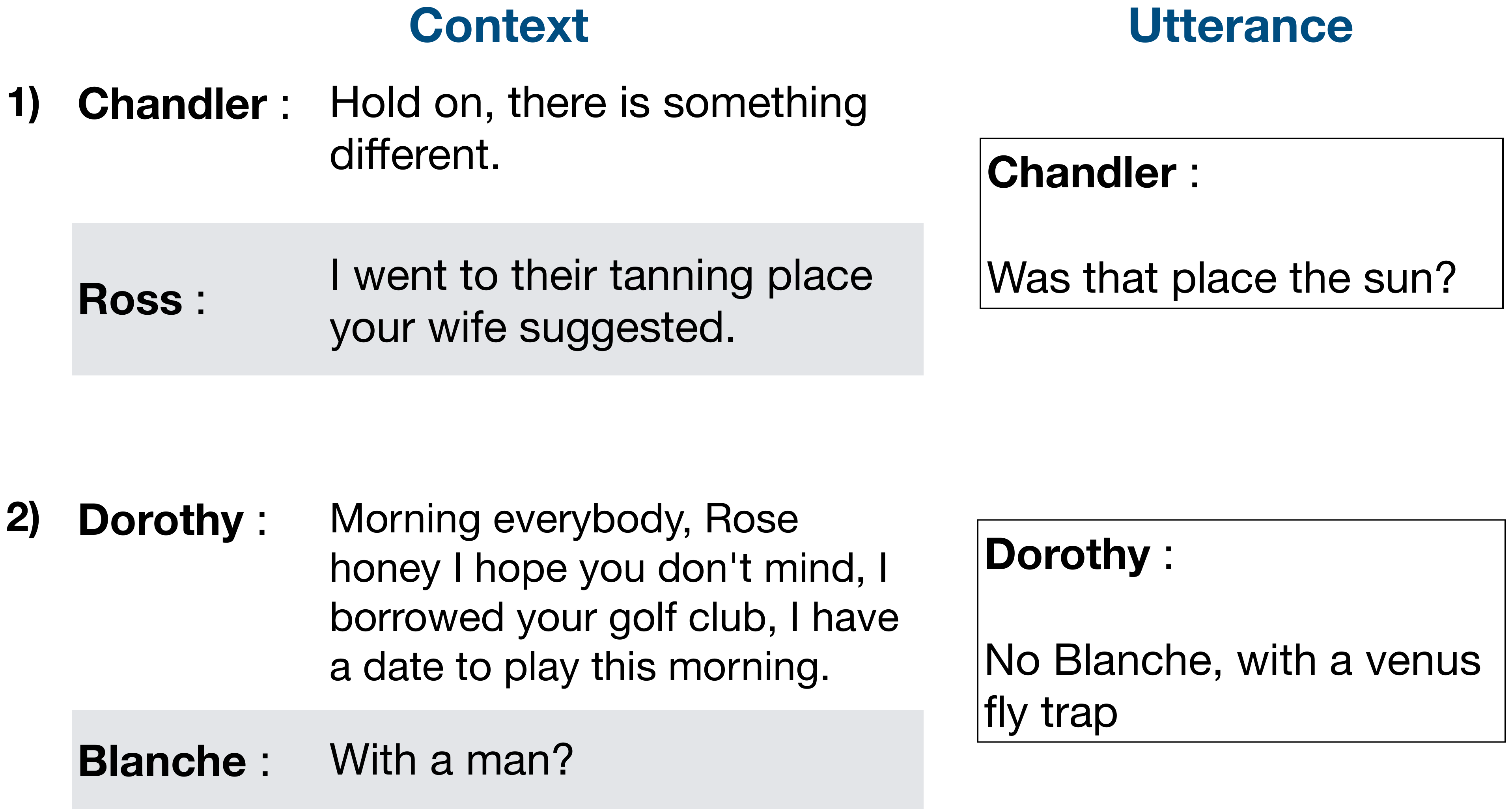}
	\caption{Context importance in sarcasm detection.}
	\label{fig:context_examples}
\end{figure}

In \cref{fig:context_examples}, we present two instances from the dataset where the role of conversational context is essential in determining the sarcastic nature of an utterance. In the first case, the sarcastic reference of the \textit{sun} is apparent only when the topic of discussion is known, i.e., \textit{tanning}. In the second case, the reference made by the speaker regarding a \textit{venus flytrap} can be recognized as sarcastic only when it is known to be referred as a thing to go on a \textit{date} with. These examples demonstrate the importance of having contextual information. The availability of context in our proposed dataset provides models with the ability to utilize additional information while reasoning about sarcasm. Enhanced techniques would require commonsense reasoning to understand illogical statements (such as going on a date with a \textit{venus flytrap)}, which indicate the presence of sarcasm.

\section{Multimodal Feature Extraction} \label{sec:feature_extraction}

We obtain several learning features from the three modalities included in our dataset. The process followed to extract each of them is described below:

\paragraph{Text Features:}

We represent the textual utterances in the dataset using BERT \cite{devlin2018bert}, which provides a sentence representation \(\mathbf{u}_t \in \mathbb{R}^{d_t} \) for every utterance \(u\). In particular, we average the last four transformer layers of the first token (\texttt{[CLS]}) in the utterance -- using the BERT-Base model -- to get a unique utterance representation of size \(d_t = 768\).
We also considered averaging Common Crawl pre-trained \(300\) dimensional GloVe word vectors~\cite{pennington2014glove} for each token; however, it resulted in lower performance as compared to BERT-based features.

\paragraph{Speech Features:}

To leverage information from the audio modality, we obtain low-level features from the audio data stream for each utterance in the dataset. Through these features, we intend to provide information related to pitch, intonation, and other tonal-specific details of the speaker~\cite{tepperman2006yeah}. We utilize the popular speech-processing library Librosa \cite{brian_mcfee_2018_1342708} and perform the processing pipeline described next. First, we load the audio sample for an utterance as a time series signal with a sampling rate of 22050 Hz. Then we remove background noise from the signal by applying a heuristic vocal-extraction method.\footnote{\scriptsize{\protect\url{http://librosa.github.io/librosa/auto_examples/plot_vocal_separation.html\#sphx-glr-auto-examples-plot-vocal-separation-py}}} Finally, we segment the audio signal into $d_w$ non-overlapping windows to extract local features that include MFCC, melspectogram, spectral centroid and their associated temporal derivatives (delta). Segmentation is done to achieve a fixed length representation of the audio sources which are otherwise variable in length across the dataset. %
\begin{figure}[htbp!]
	\includegraphics{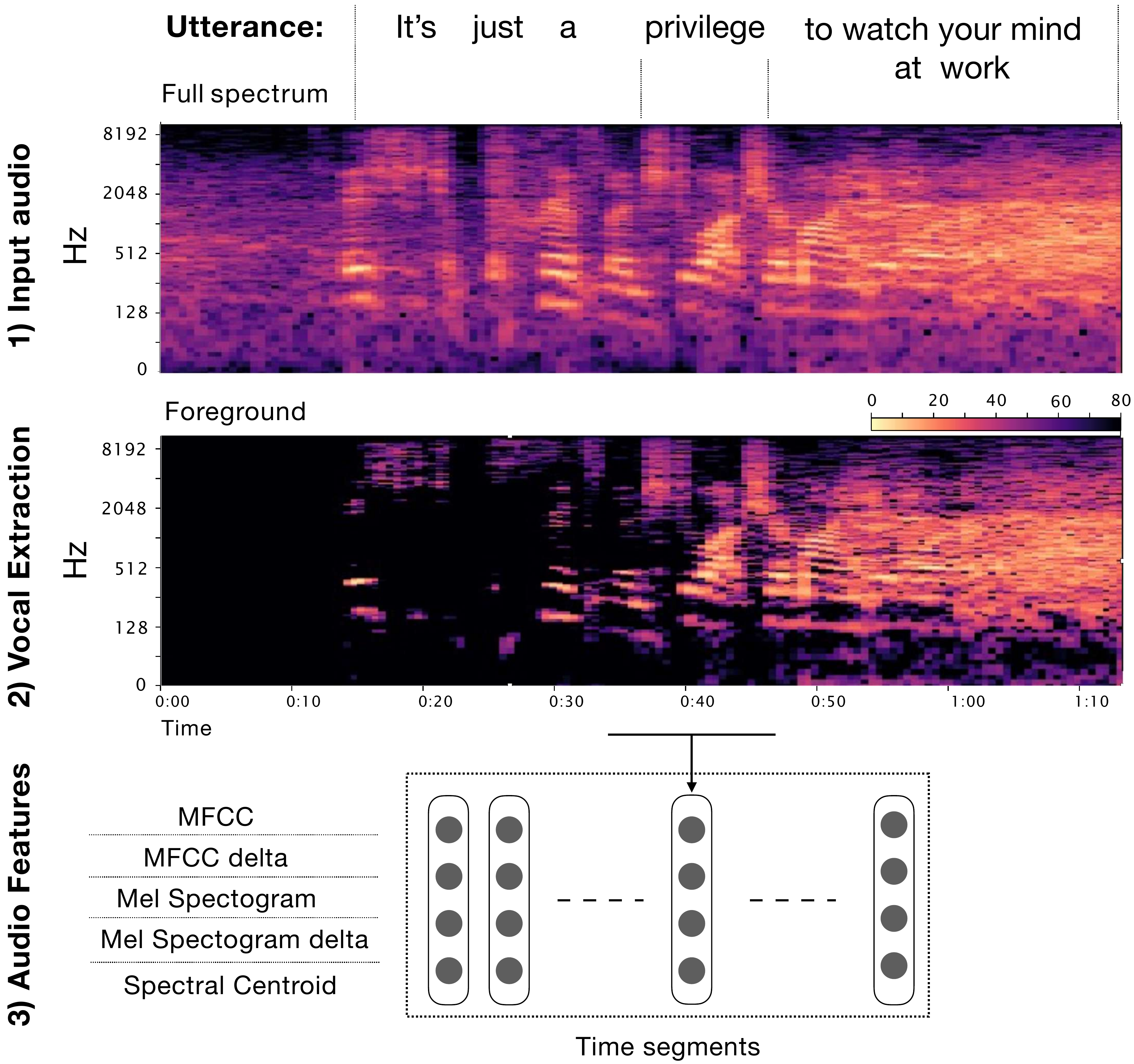}
	\caption{Feature extraction for audio modality.}
	\label{fig:audio_features}
\end{figure}
All the extracted features are concatenated together to compose a $d_a = 283$ dimensional joint representation $\{\mathbf{u}_i^a\}_{i=1}^{d_w}$ for each window. The final audio representation of each utterance is obtained by calculating the mean across the window segments, i.e.\ $\mathbf{u}_a = \frac{1}{d_w}(\sum_i\mathbf{u}_i^a) \in \mathbb{R}^{d_a}$.
\paragraph{Video Features:}
We extract visual features for each of the \(f\) frames in the utterance video using a \texttt{pool5} layer of an ImageNet \cite{deng2009imagenet} pretrained ResNet-152 \cite{he2016deep} image classification model. We first preprocess every frame by resizing, center-cropping and normalizing it. To obtain a visual representation of each utterance, we compute the mean of the obtained \(d_v = 2048\) dimensional feature vector \(\mathbf{u_i^v}\) for every frame: \(\mathbf{u}_v = \frac{1}{f}\left(\sum_i\mathbf{u_i^v}\right) \in \mathbb{R}^{d_v}\). While we could use more advanced visual encoding techniques (e.g., recurrent neural network encoding techniques), we decide to use the same averaging strategy as with the other modalities.  %

\section{Experiments}
\label{sec:experiments}
To explore the role of multimodality in sarcasm detection, we conduct multiple experiments evaluating each modality separately and also combinations of modalities provided in the dataset. Additionally, we investigate the role of context and speaker information for improving predictions.

\subsection{Experimental Setup}
We perform two main sets of evaluations. The first set involves conducting five-fold cross-validation experiments where the folds are randomly created in a stratified manner. This is done to ensure label balance across folds. In each of the $K$ iterations, the $k^{th}$ fold acts as a testing set while the remaining are used for training. Validation folds can be obtained from a part of the training folds. As the folds are created in a randomized manner, there is overlap between speakers across training and testing sets, thus resulting in a \textit{speaker-dependent} setup. The second set of evaluations restrict the inclusion of utterances from the same speaker to be either in the training or testing sets. Utterances from The Big Bang Theory, The Golden Girls and Sarcasmaholics Anonymous are made part of the training set while Friends is used as a testing set.\footnote{Split details are released along with the dataset for consistent comparison by future works.} We call this the \textit{speaker-independent} setup. Motivation for such a setup is discussed in Section~\ref{sec:multimodal_prediction}.

During our experiments, we use precision, recall, and F-score as the main evaluation metrics, weighted across both sarcastic and non-sarcastic classes. The weights are obtained based on the class ratios. For speaker-dependent scenraio, we report results by averaging across the five-fold cross-validation results. %

\subsection{Baselines}

The experiments are conducted using three main baseline methods:

\paragraph{Majority:} This baseline assigns all the instances to the majority class, i.e., non-sarcastic.

\paragraph{Random:} This baseline makes random/chance predictions sampled uniformly across the test set.

\paragraph{SVM:} We use Support Vector Machines (SVM) as the primary baseline for our experiments. SVMs are strong predictors for small-sized datasets and at times outperform neural counterparts \cite{byvatov2003comparison}. We use the SVM classifiers from scikit-learn \cite{scikit-learn} with an RBF kernel and a scaled gamma. The penalty term $C$ is kept as a hyper-parameter which we tune based on each experiment (we choose between 1, 10, 30, 500, and 1000). For the speaker dependent setup we scale the features by subtracting the mean and dividing them by the standard deviation.  Multiple modalities are combined using  early fusion, where the features drawn from the different modalities are concatenated together.

\begin{table}[t]
	\centering
	\small
	\resizebox{\linewidth}{!}{
	\begin{tabular}{l|c|rrr}
		\small{Algorithm} & \small{Modality} & \small{Precision} & \small{Recall} & \small{F-Score} \\
		\midrule
		\midrule
		Majority & - & 25.0 & 50.0 & 33.3\\
		Random & - & 49.5 & 49.5 & 49.8 \\
		\midrule
		\multirow{7}{*}{SVM}& T & 65.1 & 64.6 & 64.6 \\
		& A & 65.9 & 64.6 & 64.6 \\
		& V & 68.1 & 67.4 & 67.4 \\ \cline{2-5}
		& T+A & 66.6 & 66.2 & 66.2 \\
		& T+V & \textbf{72.0} & \textbf{71.6} & \textbf{71.6} \\
		& A+V & 66.2 & 65.7 & 65.7 \\
		& T+A+V & 71.9  & 71.4 & 71.5 \\
		\midrule
		$\Delta_{multi-unimodal}$ &  & \textcolor{green}{$\uparrow 3.9\%$} & \textcolor{green}{$\uparrow 4.2\%$} & \textcolor{green}{$\uparrow 4.2\%$} \\
		\small{Error rate reduction} &  & \textcolor{green}{\(\uparrow 12.2\%\)} & \textcolor{green}{\(\uparrow 12.9\%\)} & \textcolor{green}{\(\uparrow 12.9\%\)} \\
		\bottomrule
	\end{tabular}
	}
	\caption{Speaker-dependent setup. All results are averaged across five folds where each fold present weighted F-score across both sarcastic and non-sarcastic classes.}
	\label{tab:speaker_dependent}
\end{table}

\begin{table}[hbtp!]
	\centering
	\small
	\resizebox{\linewidth}{!}{
	\begin{tabular}{l|c|rrr}
		\small{Algorithm} & \small{Modality} & \small{Precision} & \small{Recall} & \small{F-Score} \\
		\midrule
		\midrule
		Majority & - & 32.8 & 57.3 & 41.7 \\
		Random & - & 51.1 & 50.2 & 50.4 \\
		\midrule
		\multirow{7}{*}{SVM} & T & 60.9 & 59.6 & 59.8 \\
		& A & \textbf{65.1} & 62.6 & 62.7 \\
		& V & 54.9 & 53.4 & 53.6 \\ \cline{2-5}
		& T+A & 64.7 & \textbf{62.9} & \textbf{63.1} \\
		& T+V & 62.2 & 61.5 & 61.7 \\
		& A+V & 64.1 & 61.8 & 61.9 \\
		& T+A+V & 64.3 & 62.6 & 62.8 \\
		\midrule
		$\Delta_{multi-unimodal}$ & & \textcolor{red}{$\downarrow 0.4\%$} & \textcolor{green}{$\uparrow 0.3\%$} & \textcolor{green}{$\uparrow 0.4\%$} \\
		\small{Error rate reduction} & & \textcolor{red}{\(\downarrow 1.1\%\)} & \textcolor{green}{\(\uparrow 0.8\%\)} & \textcolor{green}{\(\uparrow 1.1\%\)} \\
		\bottomrule
	\end{tabular}
	}
	\caption{Multimodal sarcasm classification. Evaluated using an speaker-independent setup. Note: T=text, A=audio, V=video.}
	\label{tab:speaker_independent}
\end{table}

\section{Multimodal Sarcasm Classification}
\label{sec:multimodal_prediction}

\cref{tab:speaker_dependent} presents the classification results for sarcasm prediction in the speaker-dependent setup. The lowest performance is obtained with the Majority baseline which achieves $33.3\%$ weighted F-score ($66.7\%$ F-score for non-sarcastic class and $0\%$ for sarcastic). The pre-trained features for the visual modality provide the best performance among the unimodal variants. The addition of textual features through concatenation improves the unimodal baseline and achieves the best performance. The tri-modal variant is unable to achieve the best score due to a slightly sub-optimal performance from the audio modality. Overall, the combination of visual and textual signals significantly improves over the unimodal variants, with a relative error rate reduction of up to 12.9\%. 

\begin{figure}[hbtp!]
\centering
	\includegraphics[width=\linewidth]{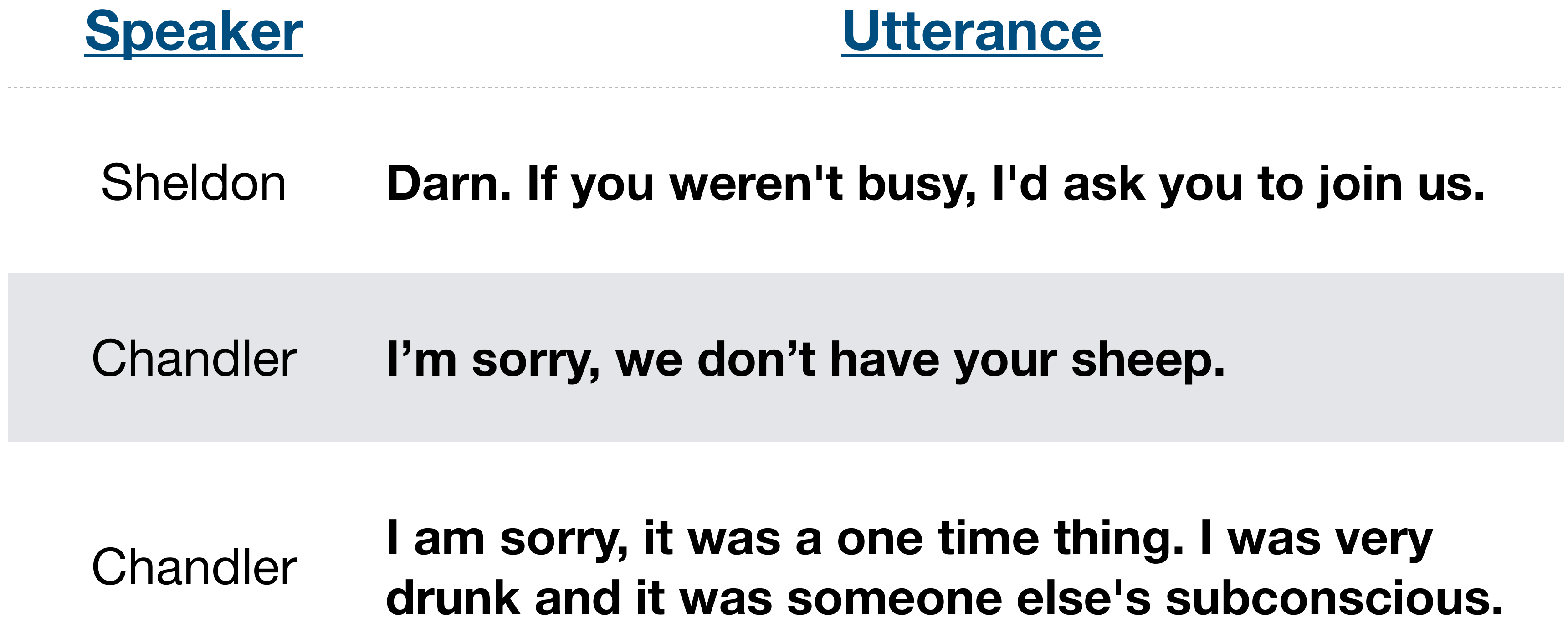}
	\caption{Sample sarcastic utterances correctly predicted by T+V but not only T model in the speaker-dependent setup. The utterances either do not have explicit sarcastic markers or need commonsense-reasoning to detect ironies, such as \textit{being drunk in someone else's subconscious}.}
	\label{fig:TVvsT}
\end{figure}

We manually investigate the utterances where the bimodal textual and visual model predicts sarcasm correctly while the unimodal textual model fails. In most of these samples, the textual component does not reveal any explicit sarcasm (see~\cref{fig:TVvsT}). As a result, the utterances require additional cues, which it avails from the multimodal signals.

The speaker-independent setup is more challenging as compared to the speaker-dependent scenario, as it prevents the model from registering speaker-specific patterns. The presence of new speakers in the testing set requires a higher degree of generalization from the model. Our setup also segregates at the source level, thus the testing involves an entirely new environment concerning all the modalities. We believe that the speaker-independent setup is a strong test-bed for multimodal sarcasm research. The increased difficulty of this task is also noticed in the model training, which now requires a smaller error margin (or higher $C$ value) of the SVM's decision function to provide good test performance.

\cref{tab:speaker_independent} presents the performance of our baselines in the speaker-independent setup.
In this case, the multimodal variants do not greatly outperform the unimodal counterparts.
Unlike \cref{tab:speaker_dependent}, the audio channel plays a more important role, and it is slightly improved by adding text.
By inspecting the correctly predicted sarcastic examples by text plus audio but not by text, we observe a tendency of higher mean pitch (mean fundamental frequency) with respect to those incorrectly predicted, as \citet{attardo2003multimodal} suggested.
Failure cases seem to contain particular patterns of high pitch, also studied by \citet{attardo2003multimodal}, but in average they seem to have normal pitch.
In this sense, future work can focus on analyzing the temporal localities of the audio channel.

In this setup, video features do not seem to work well. We hypothesize that, because the visual features are about object features (not specific to sarcasm) and the model is shallow, these features may make the model capture character biases which make them unsuitable for the speaker-independent setup. This is also suggested by the statistics in  \cref{fig:correct_prediction} which we describe in the next section.
By looking at the incorrect predictions by the best model, we infer that models should better capture the mismatches between the main speaker facial expressions and the emotions of what is being said.

\subsection{The Role of Context and Speaker Information}
\label{sec:role_of_context_and_speaker}

We investigate whether additional information, such as an utterance's context (i.e., the preceding utterances, cf\@. \cref{sec:qualitative}) and the speaker identification, are helpful for the predictions. Context features are generated by averaging the representations of the utterances (as per \cref{sec:feature_extraction}) present in the context. For the speakers, we use a one-hot encoding vector with size equal to the total unique speakers in a training fold.

\begin{table}[t]
	\centering
	\resizebox{\linewidth}{!}{
	\begin{tabular}{cl|rrr}
		\small{Setup} & \small{Features} &\small{Precision} & \small{Recall} & \small{F-Score} \\
		\midrule
		\midrule
		\multirow{6}{*}{\shortstack{Speaker\\Dependent}} & T & 65.1 & 64.6 & 64.6 \\
		 &  + context & 65.5 & 65.1 & 65.0 \\
		 &  + speaker & \textbf{67.7} & \textbf{67.2} & \textbf{67.3} \\ \cline{2-5}
		 & Best (T + V) & 72.0 & 71.6 & \textbf{71.8} \\
		 &  + context & 71.9 & 71.4 & 71.5 \\
		 &  + speaker & \textbf{72.1} & \textbf{71.7} & \textbf{71.8} \\
		\midrule
		\multirow{6}{*}{\shortstack{Speaker\\Independent}} & T & \textbf{60.9} & 59.6 & 59.8 \\
		 &  + context & 57.9 & 54.5 & 54.1 \\
		 &  + speaker & 60.7 & \textbf{60.7} & \textbf{60.7} \\ \cline{2-5}
		 & Best (T + A) & 64.7 & \textbf{62.9} & \textbf{63.1} \\
		 &  + context & \textbf{65.2} & \textbf{62.9} & 63.0 \\
		 &  + speaker & 64.7 & \textbf{62.9} & \textbf{63.1} \\ \midrule
	\end{tabular}
	}
	\caption{Role of context and utterance's speaker. Note: T=text, A=audio, V=video.}
	\label{tab:additional_info}
\end{table}

\cref{tab:additional_info} shows the results for both evaluation settings for the textual baseline and the best multimodal variant. For the context features, we see a slight improvement in the best variant of the speaker independent setup (text plus audio); however, in other models, there is no improvement. A possible reason could be the loss of temporal information when pooling across the conversation.

\begin{figure}[t]
	\includegraphics{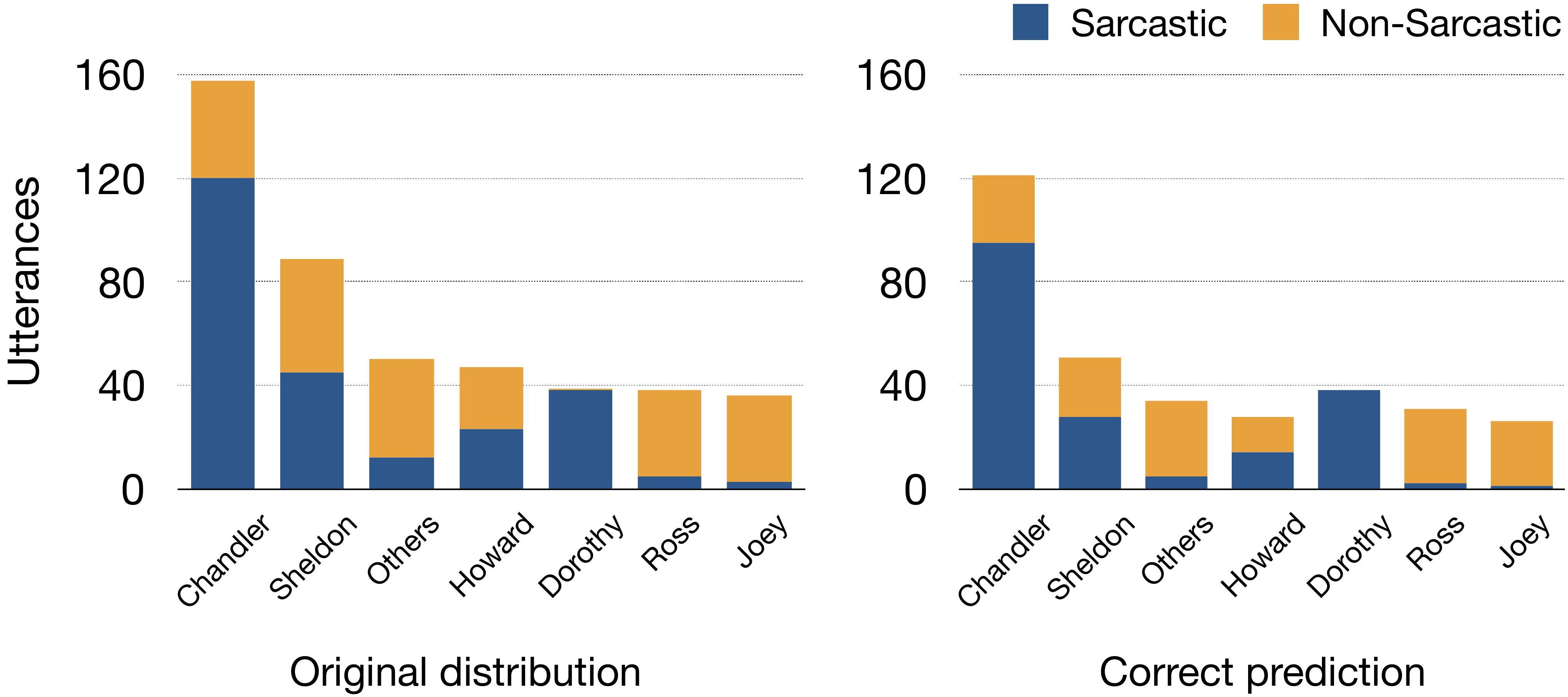}
	\caption{Correlation in speaker-specific sarcastic tendencies of the top-7 speakers. Predictions are obtained from the best performing model from \cref{tab:speaker_dependent}. Speaker identifier features are not used.}
	\label{fig:correct_prediction}
\end{figure}

For the speaker features, we see an improvement in the speaker-dependent setup for the textual modality. Due to the speaker overlap across splits, the model can leverage speaker regularities for sarcastic tendencies. However, we do not observe the same trend for the best multimodal variant (text + video) where the score barely improves. To understand this result, we visualize the correct predictions made by this model. The results, as seen in \cref{fig:correct_prediction}, show a correlation between the class distributions among the overall ground truth and the correctly predicted instances per speaker. As this model does not use speaker information, this correlation indicates that the multimodal variant is able to learn speaker-specific information transitively through the input features, rendering additional speaker input redundant. Lastly, in the speaker independent setup, the speaker information does not lead to improvement. This is also expected as there is no speaker overlap between the splits.

\section{Conclusion and Future Work}
\label{sec:conclusion}

In this paper, we provided a systematic introduction to multimodal learning for sarcasm detection. To enable research on this topic, we introduced a novel dataset, MUStARD, consisting of sarcastic and non-sarcastic videos drawn from different sources. By showing multiple examples from our curated dataset, we demonstrate the need for multimodal learning for sarcasm detection. Consequently, we developed models that leverage three different modalities, including text, speech, and visual signals. We also experimented with the integration of context and speaker information as additional input for our models.

The results of the baseline experiments supported the hypothesis that multimodality is important for sarcasm detection. In multiple evaluations, the multimodal variants were shown to significantly outperform their unimodal counterparts, with relative error rate reductions of up to 12.9\%. %

Moreover, while conducting this research, we identified several challenges that we believe are important to address in future research work on multimodal sarcasm detection.

\textbf{Multimodal fusion:} So far, we have only explored early fusion for multimodal classification. Future work could investigate advanced spatiotemporal fusion strategies (e.g., Tensor-Fusion~\cite{zadeh2017tensor}, CCA~\cite{hotelling1936relations}) to better encode the correspondence between modalities. Another direction could be to create fusion strategies that can better model incongruity among modalities to identify sarcasm.

\textbf{Multiparty conversation:} The dialogues represented in our dataset are often multi-party conversations. Advanced techniques to learn multimodal relationships could incorporate better relationship modeling~\cite{majumder2018dialoguernn}, and exploit models that provide gesture, facial and pose information about the people in the scene \cite{cao2018openpose}.

\textbf{Neural baselines:} As we strove to create a high-quality dataset with rich annotations, we had to trade-off  corpus size. Moreover, the occurrence of sarcastic utterances itself is scanty. To focus on effects induced by multimodal experiments, we chose a balanced version of the dataset with a limited size. This, however, arises the problem of over-fitting in complex neural models. 
As a consequence, in our initial experiments, we noticed that SVM classifiers perform better than their neural counterparts, such as CNNs. Future work should try to overcome this issue with solutions involving pre-training, transfer learning, domain adaption, or low-parameter models.

\textbf{Sarcasm detection in conversational context:} Our proposed MUStARD is inherently a dialogue level dataset where we aim to classify the last utterance in the dialogue. In a dialogue, to classify an utterance at time $t$, the preceding utterances at time $< t$ can be considered as its context. In this work, although we utilize conversational context, we ignore modeling various key conversation specific factors such as interlocutors' goals, intents, dependency, etc.~\cite{poria2019emotion}. Considering these factors can improve context modeling necessary for \textit{sarcasm detection in conversational context}. Future work should try to leverage these factors to improve the baseline scores reported in this paper.

\textbf{Main speaker localization:} We currently extract visual features ubiquitously for each frame. As gesture and facial expressions are important features for sarcasm analysis, we believe the capability for models to identify the speakers in the multiparty videos is likely to be beneficial for the task. %

Finally, we believe the resource introduced in this paper has the potential to enable novel research in multimodal sarcasm detection. %

\section*{Acknowledgements}
We are grateful to Gautam Naik for his help in curating part of the dataset from online resources. This research was partially supported by the Singapore MOE Academic Research Fund (grant \#T1 251RES1820), by the Michigan Institute for Data Science, by the National Science Foundation (grant \#1815291), by the John Templeton Foundation (grant \#61156), and by DARPA (grant \#HR001117S0026-AIDA-FP-045). %

\bibliography{main}
\bibliographystyle{acl_natbib}

\end{document}